\documentclass{article}

\usepackage[verbose=true,letterpaper]{geometry}
\AtBeginDocument{
  \newgeometry{
    textheight=9in,
    textwidth=5.5in,
    top=1in,
    headheight=12pt,
    headsep=25pt,
    footskip=30pt
  }
}

\usepackage[utf8]{inputenc} 
\usepackage[T1]{fontenc}    
\usepackage{hyperref}       
\usepackage{url}            
\usepackage{booktabs}       
\usepackage{amsfonts}       
\usepackage{amsmath}
\usepackage{nicefrac}       
\usepackage{microtype}      

\usepackage{graphicx}   
\usepackage{subfigure} 
\usepackage{color}
\usepackage[colorinlistoftodos]{todonotes}
\usepackage[margins,adjustmargins]{trackchanges}
\usepackage{floatrow}

\title{
On the Importance of Consistency in Training Deep Neural Networks
}

\author{Chengxi Ye$^1$, Yezhou Yang$^{2}$, Cornelia Ferm\"{u}ller$^1$, Yiannis Aloimonos$^1$
\\
$^1$Computer Vision Lab, University of Maryland, \{cxy, fer, yiannis\}@umiacs.umd.edu
\\
$^2$CIDSE, Arizona State University,  yz.yang@asu.edu\\
}

\begin{document}

\maketitle

\begin{abstract}
We explain that the difficulties of training deep neural networks come from a syndrome of three consistency issues. This paper describes our efforts in their analysis and treatment. The first issue is the training speed inconsistency in different layers. We propose to address it with an intuitive, simple-to-implement, low footprint second-order method. The second issue is the scale inconsistency  between the layer inputs and the layer residuals. We explain how second-order information provides favorable convenience in removing this roadblock. The third and most challenging issue is the inconsistency in residual propagation. Based on the fundamental theorem of linear algebra, we provide a mathematical characterization of the famous vanishing gradient problem. Thus, an important design principle for future optimization and neural network design is derived. We conclude this paper with the construction of a novel contractive neural network.
\end{abstract}

\section{Introduction}

The last decade of rapid developments in the field of 
Deep Neural Networks or Deep Learning
have laid the foundation for a large amount of major advancements in artificial intelligence, ranging from speech recognition~\cite{hinton2012deep,dahl2010context}, image recognition~\cite{krizhevsky2012imagenet}, and  natural language processing~\cite{pennington2014glove,collobert2011natural}  to motor skill learning \cite{mnih2015human}. 
While various theoretical perspectives~\cite{mallat2016understanding,lin2016does,brahma2016deep} have been proposed to explain why deep learning is successful, the general consensus of the community is to attribute the success to the joint forces of straightforward neural modeling, simple learning techniques, the availability of big data and the hardware revolution in high performance computing. To date, training large scale deep neural networks is still largely an unintuitive, computationally intensive and time-consuming process. New generations of modern hardware as well as software architectural innovations have not addressed these fundamental challenges. The community strives to come up with a simple, intuitive explanation to these mysterious networks to facilitate the design of more intelligent networks and training schema. 

Among the various operations in  training  deep neural networks, performing gradient descent iterations~\cite{lecun1998gradient} is the most costly and time-consuming. Most current training algorithms 
adopt first-order methods, i.e. modifications of the steepest descent method~\cite{sutskever2013importance,kingma2014adam,duchi2011adaptive,tieleman2012lecture}, to conduct gradient decent process. Though mathematically and practically straightforward, the convergence rate of first-order methods is inherently sub-optimal for large scale non-linear regression problems, such as the training of deep neural networks. 
\begin{figure}
\centering
\includegraphics[width=2.0in]{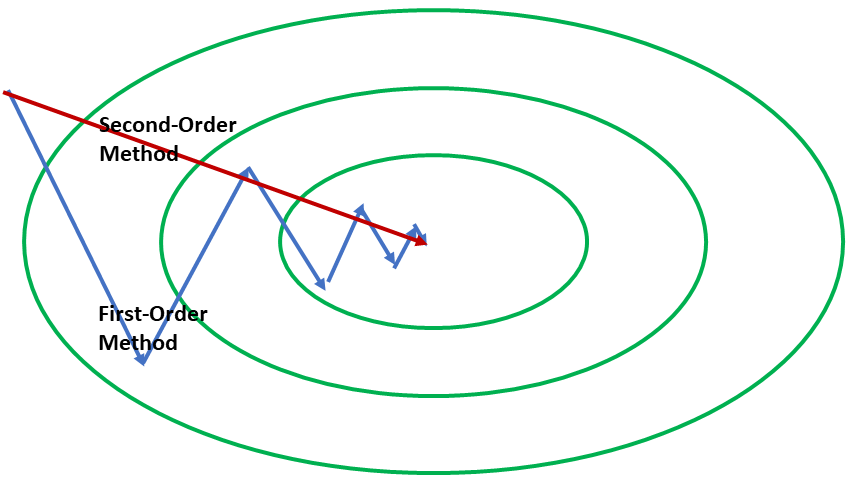}
\caption{An intuitive illustration of the first-order gradient descent method vs the second-order gradient descent method on a quadratic surface.}
\label{fig:GradientDesent}
\end{figure}
A natural extension of  first-order methods is the second-order methods (Fig.~\ref{fig:GradientDesent}), commonly known as the  Newton's method~\cite{wright1999numerical}. Intuitively speaking, unlike the near-sighted first-order methods, second-order optimization algorithms  take into account the local geometry of the underlying problem. They provide a more global point of view and are exact solutions to the quadratic approximation of the energy surface. However, implementations of second-order methods on large scale neural networks are significantly more complex, require lots of memory,  and usually do not lead to better performances, both in terms of training speed and the final model quality~\cite{sutskever2013importance}. Also, during each iteration,  second-order methods are computationally more expensive than first-order ones. In the context of large scale regression with millions or billions of parameters, utilizing second-order information is hard. Direct computation of the required second-order derivatives,  that is the Hessian matrix and its inverse, is computationally intractable, and using approximations of the Hessian is still more  costly~\cite{martens2010deep,yu2011levenberg,dauphin2014identifying} compared to the simple steepest descent method. In order to keep our discussion simple, we refer the curious readers to classic textbooks, such as~\cite{wright1999numerical}, for a detailed discussion on the implementation of these methods. Due to these difficulties, in most works deep neural networks are trained with  first-order methods. 

Nevertheless, second-order methods have several favorable properties: they have super-linear convergence rate, while first-order methods~\cite{wright1999numerical} have linear convergence rate. More importantly, they have better invariance properties: the optimal step sizes are determined with the use of curvature information and as a result are usually close to $1$. 

In this work we analyze the difficulties in training deep neural networks. In our opinion, training these networks has been found technically difficult for decades because of a \textit{combination} of three challenges. All three can be explained as inconsistency in different layers of the networks. In this paper we will discuss each of these challenges and propose second-order training techniques to tackle these problems.  

In the first part of the paper, we put forward a novel method showing that second-order information can be introduced in an intuitive way and used effectively to train neural networks. We will demonstrate that the above-mentioned properties of second-order methods can be utilized to shorten the neural networks' training time, and they lead to better solutions.  This helps us to get over the training speed inconsistency issue. The proposed method can be implemented in a straightforward way by adding one line of computer code to the well known stochastic gradient descent (SGD) method. 

In the second part, we pick up the second-order point of view to initiate a mathematical analysis of critical challenges in training deep neural networks. These challenges have remained vague in the current literature. We  explain and propose ways to alleviate the scale inconsistency problem with the help of second-order methods. 

In the third part, we present a mathematical characterization of the vanishing gradient problem in the language of linear algebra, specifically using the \textit{fundamental theorem of linear algebra}. 

These three  analyses provide novel insights into the long lasting mysteries of neural networks. Solving the above challenges is  likely to lead to advancements in both neural network theory and applications. We conclude our paper with 
such a demonstration.

For the sake of clarity, we focus more on the intuition motivating the methods. A basic mathematical analysis is presented in a concise and self-contained fashion; a more thorough analysis can be found in  classic textbooks, such as  ~\cite{wright1999numerical}.    

\section{Our Approach}

\subsection{Speed Consistency in Training Neural Networks: The Law of the Minimum}
Training a  deep layered network structure is  a non-linear regression process that minimizes the final regression loss. It usually amounts to minimizing a set of \textit{gradients} 
 at the network output $y$, also known as regression \textit{residuals}: $r=\frac{\partial z}{\partial y}$. Following the chain rule, the gradients of the intermediate layers are calculated in the order from deep-to-shallow and altogether represent the steepest \textit{ascent} direction. Minimization is achieved by moving each parameter along a \textit{descent} direction. 

In  a layered structure such as the  neural network, if certain  layers are more difficult to train than  others, these layers will impede the training of the whole network. Unfortunately, this is usually the case with first-order gradient descent methods, as there is no guarantee that a universal learning rate fits all layers. Therefore, the speed of training the slowest layer(s) becomes the bottleneck of the whole training process. This phenomenon is reflected in  the well known law of the minimum~\cite{de1994liebig}, which motivates us to design an  approach that trains each individual layer at the \textit{consistent, full} speed. 

Let the final output of the network be $z$. We denote the parameters in the $i$-th layer as  $W_i$, the input as $x_i$, and the output as $y_i=W_i x_i$. In some of our derivations we will switch to capital letters to emphasize that the computation takes batched data and therefore forms a matrix. In neural networks, $z$ is usually a cost or loss value computed from the neural network output $y$. As a natural extension to the \textit{model residual}, the \textit{residual} in layer $i$ is defined to be the \textit{gradient} $r_i(W_i)=\frac{\partial z}{\partial y_i}$. Our work adopts a local linear model: given a displacement $P$, the new residual is approximated as $r_i(W_i+P)\approx P x_i + r_i(W_i)$. Thus, the optimization problem for \textit{each layer} is formulated as: 
\begin{equation}
\min_P \frac{1}{2} ||PX_i + R_i||^2.
\label{layer_loss}
\end{equation}
Intuitively speaking, $PX_i$ measures the change of the output in layer $i$ if we change the weights in direction $P$. The optimization problem seeks a compensation displacement $P$ that minimizes the \textit{layer residual}, and the solution of Eq.~\ref{layer_loss} satisfies the so-called \textit{normal equations}: 
\begin{equation}
P X_i X_i^T = -R_i X_i^T. 
\label{normal_eq}
\end{equation}
When $X_i X_i^T$ is well-conditioned and positive definite, $P$ is the descent direction that minimizes the regression loss. When dealing with ill-conditioned $X_i X_i^T$, regularizations are used to stabilize the problem. The trust region method~\cite{wright1999numerical} is a simple modification that seeks the solution of the following equation instead:
\begin{equation}
P (X_i X_i^T + \lambda I)  = - R_i X_i^T. 
\label{normal_eq2}
\end{equation}

We point out that Eq.~\ref{normal_eq2} is closely related with both the first-order method and the second-order method. 
One should note  that the right hand side of Eq.~\ref{normal_eq2}  is the \textit{negative gradient} of the \textit{parameters} in layer $i$: $\frac{\partial z}{\partial W_i}=\frac{\partial z}{\partial y_i} \frac{\partial y_i}{\partial W_i}= r_i x_i^T$. We can view the multiplication with  matrix $(X_i X_i^T+\lambda I)^{-1}$ as an inverse correction to the \textit{parameter gradients}, which adopts parts of the second-order information. The energy surface is first inversely transformed into an isotropic one, the optimal descent step size is determined based on the curvature.
Therefore, for each layer, our proposed method corrects the \textit{parameter gradients} obtained with the standard chain rule (as in back propagation), using the \textit{inverse} of matrix $(X_i X_i^T+\lambda I)$. Putting things together, we achieve a solution that has the form of the second-order method:   
\begin{equation}
P= - \frac{\partial z}{\partial W_i}(X_i X_i^T + \lambda I)^{-1}.
\label{solution}
\end{equation}

In Sec.~\ref{sec:exp} of this paper, we put forward a method to use this simple approximation and demonstrate its efficacy. 
We can also view the process as a decorrelation add-on to the standard stochastic gradient descent algorithm, which by-itself does not decorrelate the data. 


We attribute the above optimization solution to a layer-wise application of the well-established Levenberg-Marquardt Algorithm~\cite{marquardt1963algorithm,yu2011levenberg}.
It is noteworthy that previous attempts of designing  second-order algorithms~\cite{yu2011levenberg,martens2010deep,dauphin2014identifying} did not make explicit use of the layer residuals. As a result, in these formulations the Hessian matrix in each middle layer depends on Hessian matrices in deeper layers. The corresponding algorithms are complicated both mathematically and computationally. In our layer-wise exposition, each layer has its own optimization problem that is simple and intuitive: given input $x_i$ and residual $r_{i}$ in layer $i$, how can we adjust the parameters in this layer to minimize $r_i$? This can be done with gradient descent using iterations. If an $l^2$ loss is assumed, it can be done explicitly, as we proposed. This simple optimization is easily tractable and distinguishes our method from previous second-order methods. Procedures for such optimization have  been optimized in numerical linear algebra routines and are very efficient for  practical problems when the dimension of $x_i$ is on the scale of hundreds. Therefore such techniques can be used directly for  medium-sized multilayer perceptron networks. For convolutional neural networks, applying the aforementioned derivation leads to a non-trivial inverse filtering problem. We will discuss it further in Sec.~\ref{sec:dis}. Instead, we propose a simplified version by ignoring the pixel-wise correlation and just focus on the correlation among feature maps. This partial decorrelation can be seen  as an application of the preconditioning technique~\cite{wright1999numerical}. We treat each \textit{feature map} as a \textit{hidden unit} and calculate the correlation across different feature maps. Inverse correction is then applied to the convolution kernel gradients, as shown in Eq.~\ref{solution}. 


There are two major computational bottlenecks in this method: 1) the matrix multiplication of $X_iX_i^T$: in convolutional networks, $X_i$ is usually a highly redundant wide matrix. In our experiments, we  can significantly accelerate the matrix multiplications with no noticeable loss of accuracy  by performing a uniform subsampling. The computational cost for this is negligible compared to the whole training process.  2) The matrix inversions: this can be performed reasonably fast only at the small scale of hundreds of entries. For larger layers that contain thousands of hidden units, matrix multiplications and inversions are extremely slow. Thus, we propose to speed  up the decorrelation process using  a simple stochastic preconditioning strategy. In each training iteration, the hidden units are stochastically divided into chunks (of hundreds) so that matrix multiplications and inversions can be calculated quickly. Empirically we observe that this stochastic division and preconditioning technique is an effective way of conducting acceleration.

\subsection{Scale Consistency in Training Deep Neural Networks}

We notice that even after the removal of the computational bottleneck, direct application of the above algorithm to 10-layer neural networks leads to training failures. An analysis of the individual layer inputs and layer residuals leads us to the following observation: in the first layer of the network, inputs may be in the range of $[-1,1]$ and the residuals  in the range of $[-0.1,0.1]$, which is well-posed. In a deeper layer, the inputs may be in the range of $[-0.001,0.001]$ and the residuals in the range of $[-10,10]$. For the latter,  the optimization becomes much harder, affecting any method, including gradient descent and $l^2$ optimization. This observation reveals a second roadblock in training neural networks: scale inconsistency.

Note that a neural network is an ordered set of transforms into different spaces. During the transform, there is no constraint guaranteeing that  the input signal $x_i$ in layer $i$ will be  on a meaningful scale with  residual $r_i$. This scale inconsistency makes simple optimization attempts futile. One  approach to rectify this issue is to use normalization techniques so that the inputs of each layer are approximately on the same scale. The widely-adopted batch normalization algorithm~\cite{ioffe2015batch} was designed specifically for first-order methods. Note that a globally-optimal learning rate is almost impossible to find with first-order methods in a deep layered structure. The batch normalization algorithm introduces an extra pair of parameters, which scales and shifts the normalized results to alleviate this problem. On the other hand, for second-order methods, this pair of parameters is redundant as the scaling and shifting are  figured out automatically by the normal equations (Eq.~\ref{normal_eq2},~\ref{solution}). Speed is consistently optimal in each layer as long as the inputs are approximately on the same scale. We also observe that the rescaling in batch normalization potentially leads to model instability if the scaling factor is larger than $1$. For these two reasons we adopt a minimal normalization strategy and divide the inputs by their smoothed \textit{root mean square}: $RMS=\sqrt{\frac{1}{n}\sum_{i=1}^{n} x_i^2}$. $RMS$ can be estimated from minibatches and smoothed using moving average. On the other hand, this $RMS$ normalization is also necessary for second-order methods so that the regularization in Eq.~\ref{normal_eq2} takes \textit{consistent effects} in each layer. That is, $\lambda=0.1$ is appropriate if the inputs are in the range of $[-1,1]$, but very likely ineffective if the inputs are in $[-0.001,0.001]$. In the latter case, the displacement $P$ in Eqs. ~\ref{normal_eq2}, ~\ref{solution} is likely to explode. 

According our extensive tests, we notice that with these specific designs the previous difficulties in applying second-order methods in deep learning are largely reduced. As demonstrated in later sections (~\ref{sec:deep_nets},~\ref{sec:contractive_exp}), this simple design of an $RMS$ based normalization not only allows using second-order algorithms to train deep neural networks, but leads to the construction of an important class of neural networks.

\subsection{Clarification of Vanishing Gradient with Operator Analysis}

After normalizing the inputs into well-defined ranges, we observe that it usually becomes possible to use second-order information for training networks of $10-20$ layers. However, we encounter failures and noticeabe worse performance when the network goes beyond this depth. This observation motivates us to conduct a further mathematical analysis through operator theory~\cite{kreyszig1989introductory}.

Formally speaking, let $\mathcal{T}$ be the operator that is `the best linear approximate' of \textit{one or several} layers of the \textit{forward} transform. 
In terms of differential geometry, $\mathcal{T}$ is the \textit{differential} of the forward transform~\cite{manfredo1976carmo}.
In its `well-known' natural state~\cite{szegedy2013intriguing}, $\mathcal{T}$ is usually expansive or non-contractive, meaning that it has singular value(s) larger than $1$. This property can be intuitively understood for certain parts of the network, such as for example the normalization step after $ReLU$.

Using the notation of operator theory, $\mathcal{T}$ is an operator that maps Hilbert space $H_1$ to $H_2$. As we will show shortly, its Hilbert \textit{adjoint} $\mathcal{T}^*$ is used in the \textit{back propagation process} and maps $H_2$ back to $H_1$. According to operator theory, $\mathcal{T}^*$ has the same characteristics as $\mathcal{T}$, and therefore it should also be \textit{expansive}~\cite{kreyszig1989introductory}.
However, our investigation into the gradient propagation has a \textit{contrary} observation: a significant energy \textit{decay} is often observed when we move to the shallower layers. Empirically, this \textit{contractive} effect is known as the vanishing gradient problem in the training of deep neural networks. This problem has haunted the neural network community for decades~\cite{hochreiter2001gradient,hochreiter1997long}, even though our operator $\mathcal{T}$, therefore $\mathcal{T}^*$, is \textit{non-contractive}. The mystery of the vanishing gradient problem can be unveiled by investigating the adjoint operator $\mathcal{T}^*$. Let the forward transform be:  
\begin{equation}
z=f(y)= f(\mathcal{T} (x) )
\label{forward}
\end{equation}
With a slight abuse of notation, following the chain rule:
\begin{equation}
\frac{\partial z}{\partial x}=\frac{\partial z}{\partial y}\frac{\partial y}{\partial x}= \mathcal{T}^* \frac{\partial z}{\partial y}.
\label{backward}
\end{equation}
Therefore if  $\frac{\partial z}{\partial y}$ falls into the \textit{kernel} of $\mathcal{T}^*$ (i.e., the nullspace), then the gradient propagation \textit{stops}. In other words, in order to avoid the vanishing gradient problem, we should enforce the constraint that:
\begin{equation}
\frac{\partial z}{\partial y} \notin ker(\mathcal{T}^*)=im(\mathcal{T})^{\perp},
\label{constraint}
\end{equation}
with $\perp$ denoting the orthogonal complement.

In the real space we have $\mathcal{T}^*=\mathcal{T}^T$, and  the last equality simply follows from the  the \textit{fundamental theorem of linear algebra}~\cite{strang1993fundamental}.

Let us emphasize that $\frac{\partial z}{\partial y} \neq 0, \frac{\partial z}{\partial y} \in im(\mathcal{T})$  describes a critical constraint that should be enforced in training deep neural networks. Unsurprisingly, most of the current techniques enabling a training of very deep networks exploit this constraint. For example, the use of $ReLU$~\cite{krizhevsky2012imagenet} instead of other activation functions; the introduction of identity maps~\cite{hochreiter1997long,he2016deep}; or
the procedure of adding random noise to the gradients~\cite{neelakantan2015adding}, all help lessen the vanishing gradient problem.

In our experiments, we tested the procedure of injecting noise~\cite{neelakantan2015adding}, which showed promising results in alleviating the vanishing gradient problem. Though we have not always been successful, potentially due to the expansive nature of the transform. An intuitive explanation for why noise injection works is as follows: the optimization process stochastically perturbs the gradient in order to \textit{avoid} that the gradients fall into the kernel space, where $P$ now becomes: 
\begin{equation}
P= - (\frac{\partial z}{\partial W_i}+\epsilon)(X_i X_i^T + \lambda I)^{-1}.
\label{solution2}
\end{equation}
Here, small random noise is a valuable ingredient that adds a factor of random exploration in the training. If the perturbed gradient direction is close to a better solution, our approach makes use of the second-order information to follow it to a better solution.

Furthermore,  the analysis suggests that the $ReLU$ operator is not the best option for  gradient propagation. Here, we propose to investigate the modulus operator, which has a solid mathematical foundation in the field of wavelets analysis~\cite{mallat2016understanding}. Specifically, in order to maintain a large \textit{image space} or equivalently to shrink the \textit{kernel space}, we adopt the modulus unit $ModU$ as activation function instead of $ReLU$. In the real number space the modulus unit computes the absolute value of its input:
\begin{equation}
ModU(x)=
    \begin{cases}
      x, & \text{if}\ x>=0 \\
      -x, & \text{if}\ x<0
    \end{cases}
\label{modu}
\end{equation}

It is well known that current neural networks are susceptible to these `adversarial attacks'~\cite{szegedy2013intriguing}. It is an open problem how the existing powerful networks can be modified so that they become stable. With out stability, important applications of the neural networks such as object classification and biometric verification are exposed to clever hacks. Experiments conducted in Sec.~\ref{sec:exp}, validate that this choice of activation function is promising upon this difficult problem.

\section{Experiments}
\label{sec:exp}
The  focus of the experiments section is on demonstrations and applications of the favorable properties of second-order methods.
Second-order methods are relatively uncommon in the deep learning community. As currently we do not have competitive implementations of existing  second-order methods available in our test platform, we will compare
our proposed second-order modification only to  state-of-the-art first order stochastic gradient descent algorithms~\cite{duchi2011adaptive,kingma2014adam,tieleman2012lecture,sutskever2013importance}. Because our method utilize second-order information, we name it SGD2. The algorithms in this paper are implemented using Matlab and the source code is available in LightNet~\cite{ye2016lightnet}, a lightweight deep learning package.

\subsection{A Small Network on the MNIST dataset}
 
\begin{figure}
\centering
\subfigure[] {\includegraphics[width=2.5in]{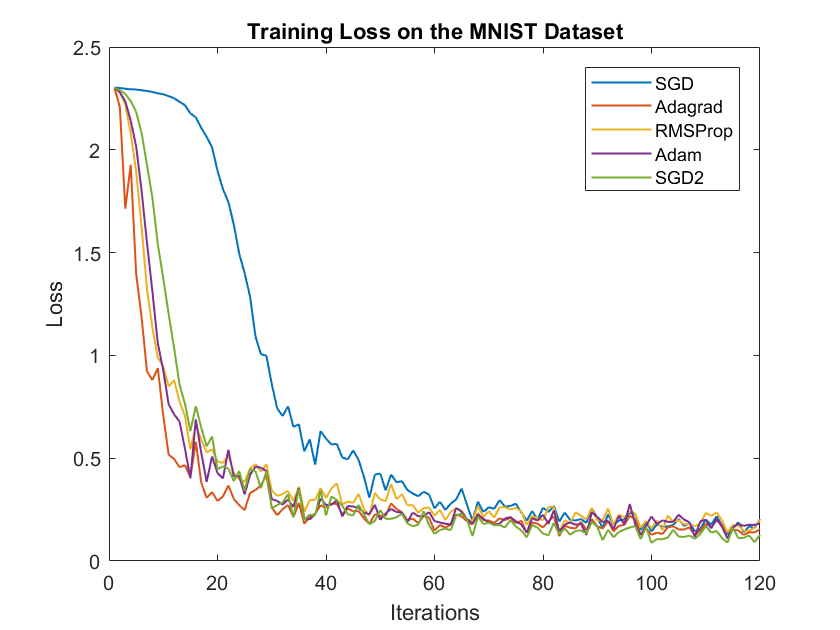}}
\subfigure[]{\includegraphics[width=2.5in]{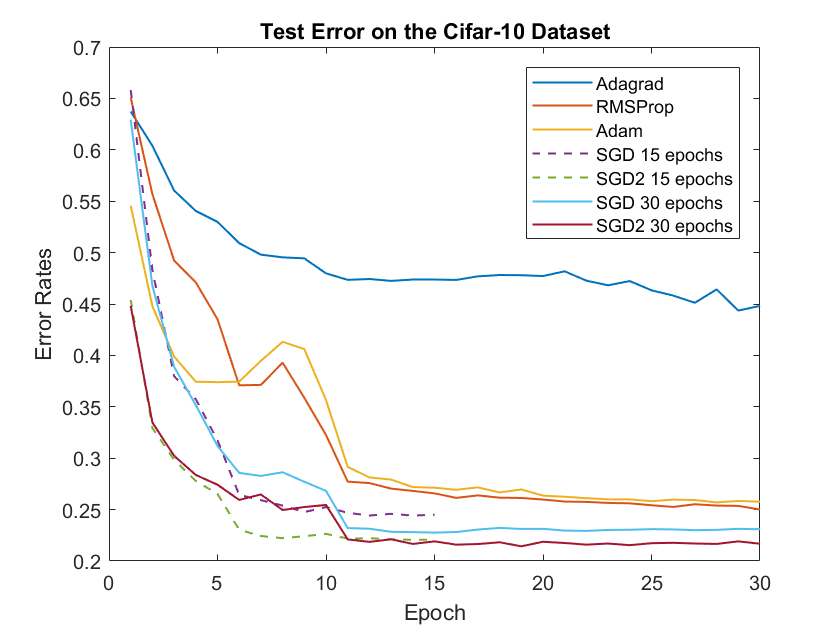}}
\subfigure[]{\includegraphics[width=2.5in]{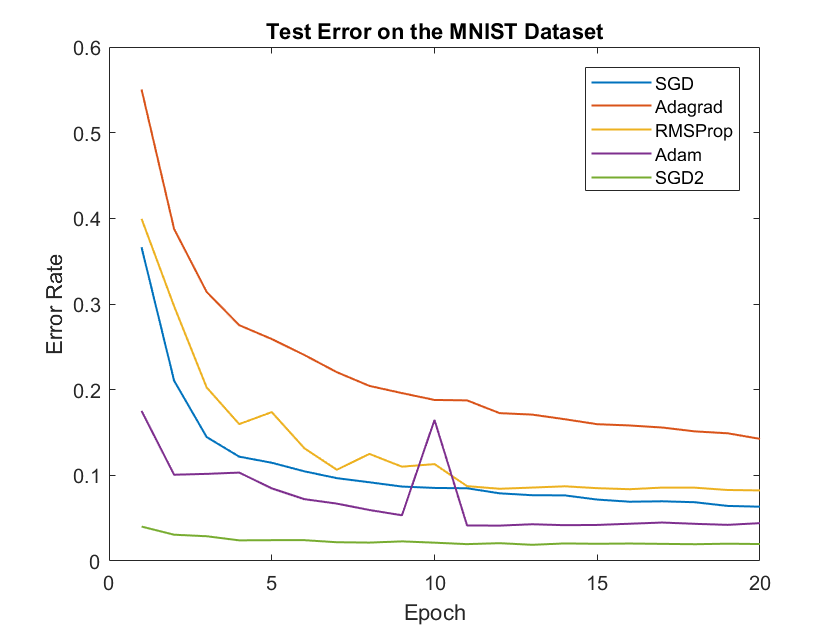}}
\subfigure[]{\includegraphics[width=2.5in]{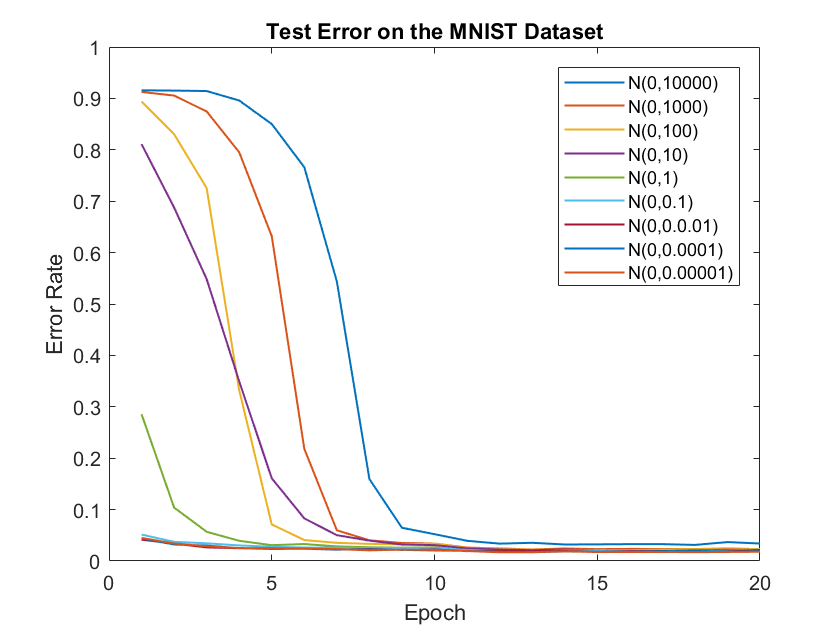}}
\caption{Training curves on the various datasets and networks.(a) Training loss of a two hidden-layer MLP network in the first epoch of the MNIST dataset. (b) Test errors on  a $5$ convolutional-layer network on the CIFAR-10 dataset. (c) Test errors of a $10$ hidden-layer MLP network on the MNIST dataset. The increases of error rates in (b,c) are caused by overfitting. (d) Test errors of a $10$ hidden-layer MLP network on the MNIST dataset. The network weights are initialized from $N(0,10^{-4})$ to $N(0,10^{4})$ and trained using second-order information.}
\label{fig:TrainingCurves}
\end{figure}

We construct a small Multilayer Perceptron network, with $2$ hidden layers. Each of these hidden layers has $128$ hidden nodes, and the network is fully-connected. $ReLU$ functions are used as the activation functions. The network weights are initialized using Gaussian random noise with a standard deviation of $0.01$. The batch size is $500$. The optimal learning rate is selected using a grid search of 30 iterations in each experiment. These settings are hold fixed in all the following experiments if not otherwise mentioned. 
In the comparison, we apply decorrelation to the gradients of each of these hidden layers, the regularization weight $\lambda$ is fixed to the value $1.0$. A comparison with state-of-the-art training algorithm demonstrates that significant acceleration is achieved by using second-order information upon its first-order counterpart (Fig.~\ref{fig:TrainingCurves}(a)). 
 
The test errors after training for the first epoch, are listed in columns $3$ of Table~\ref{ResultsTable}.
As can be seen, our method (SGD2) has significantly lower error than the standard first-order SGD (with momentum) and the methods in ~\cite{duchi2011adaptive,kingma2014adam,sutskever2013importance,tieleman2012lecture}.%

\begin{table}[]
\small
\centering
\caption{Results of various networks on MNIST~\cite{lecun1998gradient} and CIFAR-10~\cite{krizhevsky2009learning} datasets.}
\label{ResultsTable}
\begin{tabular}{|l|r|r|r|r|r|r|}
\hline
              & \multicolumn{2}{c|}{MNIST 2-layers (1 epoch)} & \multicolumn{2}{c|}{CIFAR-10 (30 epochs)} & \multicolumn{2}{c|}{MNIST 10-layers (20-epochs)} \\ \hline
Method        & Learning Rate      & Test Error     & Learning Rate   & Test Error  & Learning Rate      & Test Error     \\ \hline
SGD & 0.1                & 4.56\%         & 0.1             & 22.76\%     & 0.1                & 6.34\%         \\ \hline
Adagrad~\cite{duchi2011adaptive}      & 0.005              & 4.41\%         & 0.001           & 44.36\%     & 0.005              & 14.26\%        \\ \hline
RMSProp~\cite{tieleman2012lecture}      & 0.001              & 4.59\%         & 0.001           & 25.01\%     & 0.001              & 8.24\%         \\ \hline
Adam~\cite{kingma2014adam}        & 0.01               & 4.79\%         & 0.01            & 24.42\%     & 0.01               & 4.13\%         \\ \hline
SGD2 (ours) & 1                  & {\bf 3.33\%}         & 1               & {\bf 21.43\%}     & 1                  & {\bf 1.9\%}         \\ \hline
\end{tabular}
\end{table}

\begin{table}[]
\centering
\caption{Lowest test errors using initializations of different scales. The initialization distributions is normal distributions with standard deviations from $10^{-4}$ to $10^4$. The experiment is conducted on the MNIST dataset using the network as in section ~\ref{sec:mlp_10l}.}
\label{Initialization}
\begin{tabular}{|l|r|r|r|r|r|r|r|r|r|}
\hline
Init STD & $10^4$ & $10^3$ & $10^2$ & $10$ & $1$ & $0.1$ & $10^{-2}$ & $10^{-3}$ & $10^{-4}$ 
\\ \hline
Weight Decay & $10^{-3}$ & $10^{-3}$ & $10^{-3}$ & $5*10^{-4}$ & $5*10^{-4}$ & $10^{-4}$ & $10^{-4}$ & $10^{-4}$ & $10^{-4}$ \\ \hline
Test Error                  & 0.0313     & 0.0226    & 0.0211   & 0.02    & 0.0179 & 0.018    & 0.0179    & 0.017      & 0.0169      \\ \hline
\end{tabular}
\end{table}

\subsection{A Convolutional Network on the CIFAR-10 Dataset}

A convolutional network with $4$ convolution layers and $1$ fully-connected layer has been constructed on the CIFAR-10 dataset~\cite{krizhevsky2009learning}. This network structure has been borrowed from~\cite{vedaldi2015matconvnet}. There are $32,32,64,64$ convolution kernels of size $5\times 5$ in the first three layers, the last layer has kernel size $4\times 4$. We test the training in $30$ epochs and automatically select the best learning rate at $1/3$ and $2/3$ of the training process. To save the computational cost of matrix multiplications, we selectively apply uniform subsamplings to keep at most $10k$ data samples for each feature map. With this modification there is no noticeable overhead in calculating Eq.~\ref{solution}.

The lowest test error rates and best initial learning rates are listed in Table~\ref{ResultsTable}. We notice that on this more complex dataset our SGD2 shows distinct speed up in the first epoch, and the error rate is reduced from $65\%$ to $45\%$. The closest matching first-order method is Adam~\cite{kingma2014adam}, which has $55\%$ error rate. Scrutinizing on the training error curves (not plotted here due to space) also tells us that some of the accelerated first-order experience certain levels of overfitting near the $8$-th training epoch, results in increased error rates.For the two performers that provide better accuracy (SGD and SGD2) we tried to shorten the training epochs. We notice that fewer epochs are required to finish the training with SGD2. The final error rate is also better with second-order information (Table~\ref{ResultsTable}).

\subsection{Training a 10-layer Deep Network}
\label{sec:mlp_10l}
We straightforwardly extend the multilayer perceptron network to 10 layers. Unfortunately no training algorithm is able to make progress in the training.

We then normalize the inputs to each hidden layer by dividing their \textit{smoothed root mean square} using minibatch data and a momentum of $0.1$. Using this simple normalization scheme, the inputs to each hidden layer are on the same scale. In the back propagation step, the gradients are scaled by this normalization factor. The networks are trained in $20$ epochs and the learning rates are automatically selected then adjusted at the beginning and in the middle  of the training process using grid search.

The results are shown for the MNIST data set in Fig.~\ref{fig:TrainingCurves} (c) and the right columns of Table.~\ref{ResultsTable}. Deep networks are significantly slower and harder to train compared with shallow networks. It is non-trivial for first-order methods to select an optimal learning rate for each layer, even though the input data is properly scaled and the layer size is the same. However with second-order modification, not only the training is significantly faster, but also the final error rate is significantly lower. This observation is not unexpected as the parameter solving process is fundamentally an \textit{inverse problem}. And the optimal learning rate is related to the condition number of the different data covariance matrices in each layer used in the second-order method. Using a suboptimal learning rate, as is done with first-order methods, not only slows down the training but can result in poor local solutions. The optimal learning rate is approximately uniform thanks to the invariance properties of second-order methods. We reach from this experiment as a major conclusion  that second-order methods  can be adopted elegantly to train deep neural networks.

\subsection{Tolerance to Initializations}
In the next experiment we initialize the network weights using Gaussian distribution with standard deviations from $10^{-4}$ to $10^{4}$. Even though the initializations spans a wide scale range, direct 20-epoch trainings yield comparable final accuracies under all these settings (Fig.~\ref{fig:TrainingCurves}(d), Table ~\ref{Initialization}) . For large initializations we slightly increase weight decay regularizations to reduce overfitting and accelerate convergence. The initializations using standard deviations below $10$ lead to error rates below $2\%$. The worst test error rate of $3.13\%$ occurs when the network is badly initialized with $N(0,10^{4})$. This experiment also tells us that better convergence properties of the second-order methods can be inherited to improve training deep neural networks. The widely adopted strategy~\cite{sutskever2013importance} that neural networks need to be initialized with small random noise likely originates from a restriction of the first-order training algorithms.

\subsection{On Deeper Networks}

Since second-order methods allow us to make consistent progress in different layers of the network, we will move our focus on the second-order method to analyze the remaining challenges in training deeper networks. We explore the highly challenging task of training very deep but very narrow networks. In the following analysis, each hidden layer of a narrow neural network is limited to have only $20$ hidden nodes. 

\begin{figure}
\centering
\subfigure[] {\includegraphics[width=1.7in]{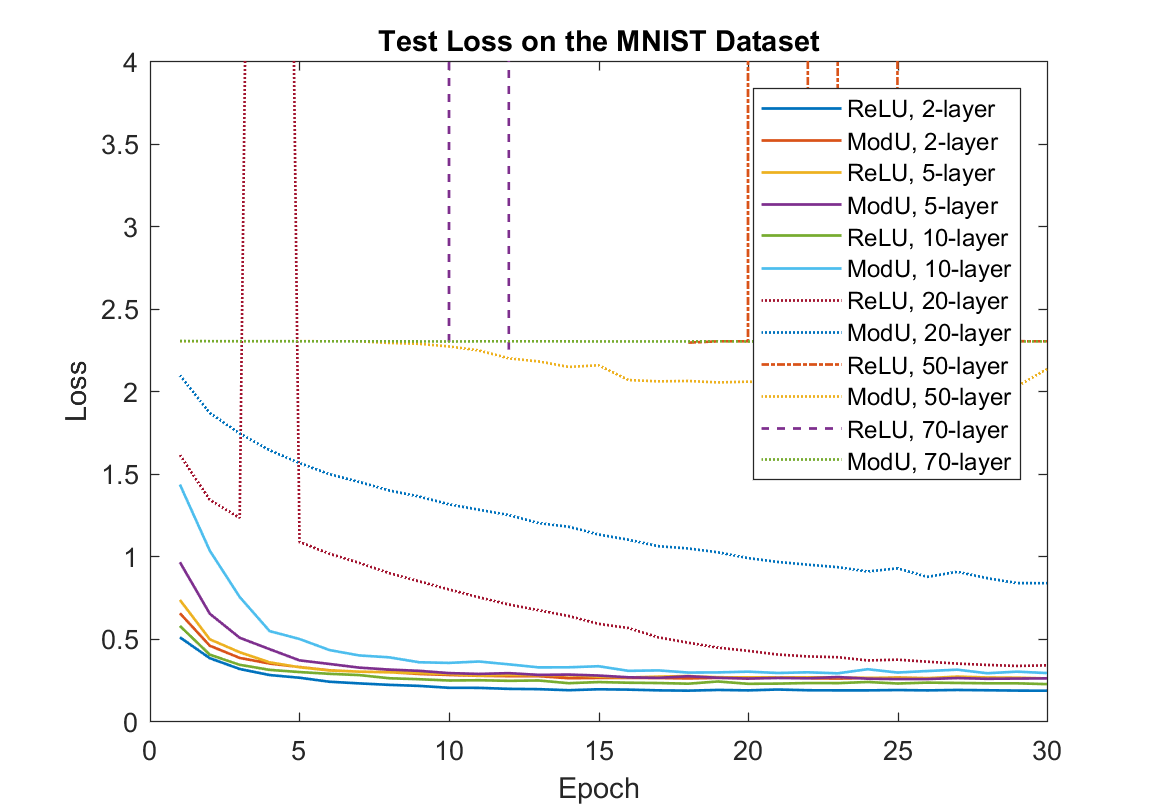}}
\subfigure[] {\includegraphics[width=1.7in]{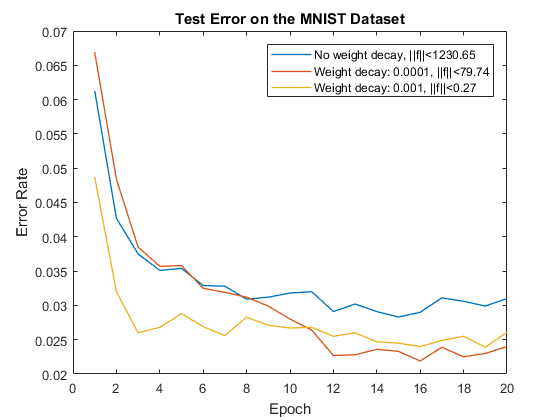}}
\subfigure[] {\includegraphics[width=1.7in]{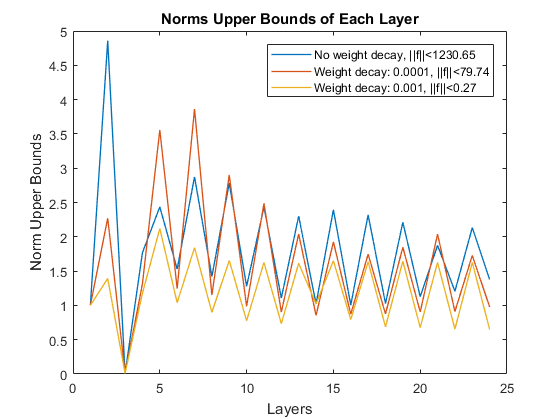}
}
\caption{Applications of $ReLU$ and $ModU$ activation functions in training deep neural networks. (a) Test losses using $ReLU$ and $ModU$ activation functions on various depth narrow networks. Each layer has $20$ hidden nodes. (b) Test error rates of a $10$-layer MLP trained with $ModU$  under different levels of weight decay regularization. Each layer has $128$ hidden nodes. (c) Norm upper bounds of (b) for linear layers interlaced by normalizations.}
\label{fig:modu}
\end{figure}

In order to create a larger image space or equivalently a smaller kernel space, we change our activation function from the popular rectified linear units $ReLU$ to modulus units $ModU$. 
Under the same settings, we are able to train a $50$-layer deep and narrow neural network with $ModU$. Its $ReLU$ counterparts lead to exploding losses (Fig.~\ref{fig:modu}). The better gradient propagation property of $ModU$ can also be observed in Fig.~\ref{fig:GradientEnergy}. However, we notice that in shallower networks the  $ModU$ function leads to slightly worse accuracy, which we believe is a main reason for  its low popularity.

\subsection{Contractive Deep Networks}
\label{sec:contractive_exp}
Nevertheless, we should not be disappointed with the somewhat mediocre results  using $ModU$. We point out that $ModU$ has another valuable property: it is non-expansive and it does not zero out its inputs. In our experiments we find that the normalization factors are mostly around $1.0$, if no or only weak regularization is applied. In contrast, if $ReLU$ is used as the activation function, we frequently observe small normalization factors in deeper layers, leading to expansive effects of the neural network. Small fabricated noise potentially leads to catastrophic effects~\cite{szegedy2013intriguing}. 

For networks with $ModU$ activation and the $RMS$ normalization as proposed here, the upper bound of the operator norm is easy to compute. We therefore calculate an upper bound by the product of the largest singular values of the linear transforms divided by the smallest normalization factors in the normalization transforms. The $ModU$ layer has norm 1.0. In our experiment on a 10-layer multilayer perceptron network, direct training using the $ModU$ on the MNIST dataset produces a norm upper bound of $1230.6$. If no regularization or weak regularization is applied, the linear transforms are likely to be expansive maps, as shown  by the bottom  zigzag curves in Fig.~\ref{fig:modu}(c). These linear layers have largest singular values above $1.0$. We then try to introduce a moderate weight decay regularization using a factor of $0.001$. The linear transform now is  regularized to be contractive, normalizations move the signal back to the right range before the next linear layer. However, the full regularized network has a  norm upper bound of $0.27$, and therefore is \textit{globally contractive} (Fig.~\ref{fig:modu}(c)). This experiment showed that neural networks can be simultaneously \textit{deep} and \textit{stable}. Unlike previous attempts which put a penalty on the Jacobian matrix ~\cite{rifai2011contractive} which is usually huge and therefore does not scale well, our constructions here are efficient architecture changes. The network can be trained using off-the-shelf training procedures.


\section{Discussions}
\label{sec:dis}
\subsection{Are Deeper Networks Better?}
\label{sec:deep_nets}

\begin{figure}
\centering
\includegraphics[width=2.5in]{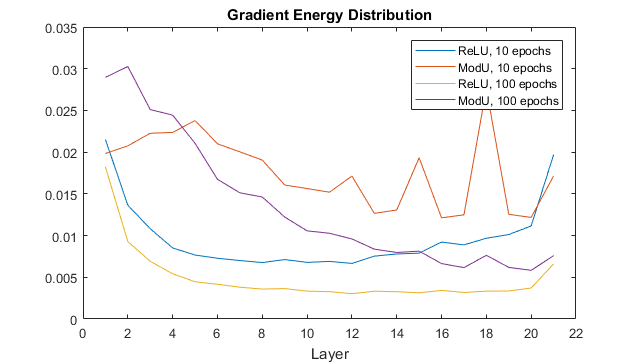}
\caption{Gradient energy distribution in a $20$-layer narrow MLP trained with $ModU$ and $ReLU$. The average energy is calculated for each layer. $ReLU$ functions have faster information loss and error contraction rates, while $ModU$ has  opposite behavior. If the network is trained sufficiently long, the error reduction will be better. Notice the curves are flatter in the middle.}
\label{fig:GradientEnergy}
\end{figure}

By using second-order training and $RMS$ normalization, we analyzed the gradient distribution in a consistent scale and \textit{training degree}. In the gradient energy distribution plot (Fig.~\ref{fig:GradientEnergy}) we see a clear and stable distribution pattern across the  different layers of the  neural networks. The gradient energy is measured by the average $l^2$ norm ($RMS$). 
Training a neural network is the process of improving the ability of \textit{error reduction}. We interpret that a mixture of two factor explains the curves in Fig.~\ref{fig:GradientEnergy}: the decaying trend from \textit{left to right} represents the \textit{error reduction} capability of neural networks. A longer training of deeper networks has a  stronger error reduction effect. Deeper networks have an accumulate effect of getting better. However, the other decaying trend, from right to left, represents the \textit{information loss} or \textit{gradient decay}. The more gradient decay, the less information is passed to the first layer of the network. This \textit{information loss} seems to set limits on the lower bound of error reduction and on the speed of improvement. The design of deep neural network is therefore a trade-off between \textit{error reduction} and \textit{information loss}. Different activation functions have different trade-offs, as can be seen from Fig.~\ref{fig:GradientEnergy}. On the other hand, different network architectures also have different trade-offs. Currently there are two major transform architectures in the community. The classic type, which goes through highly non-linear transforms, is  likely to have a faster rate in both error reduction and information loss. The other type ~\cite{hochreiter1997long,he2016deep} is more \textit{isometric}~\cite{manfredo1976carmo}, and therefore both rates will be lower. Our observations suggest that for  classic type,  activation functions and network width should provide constraints on network depth. The steep patterns at both ends and flat pattern in the middle of the curve also suggest that the \textit{error reduction} is getting \textit{slower} as the depth increase. This  gives us hints on efficient neural network design.
We conjecture these observations may have deep connections with biological neural networks.  And we expect future work to investigate these important balances in various network settings.

\subsection{Pixel-wise Decorrelation?}
Referring to Eq.~\ref{normal_eq2}, for each feature map, right multiplication with $X X^T$ represents calculating a convolution followed by a correlation. If we assume circular boundary conditions these operations can be accelerated with the Fast Fourier Transform (FFT). Pixel-wise decorrelation can be calculated as: $F(P)=\frac{-\mathcal{F} (\frac{\partial z}{\partial w}) }{\mathcal{F}^*(X) \circ \mathcal{F}^(X)+\lambda}$. Future work will investigate whether such an approach can lead to efficient implementations.

\section{Conclusion}
We made the observation that the long standing challenge of training deep artificial neural network is caused by a syndrome of three inconsistency problems. Mathematical analysis combined with tools of linear algebra, led to practical algorithms and intuitive explanations to the mysterious properties of deep neural networks. 


\small
\bibliographystyle{plain}
\bibliography{references.bib}
\end{document}